\documentclass[letterpaper]{article} 
\usepackage[preprint]{format/aaai2027} 

\usepackage[
    letterpaper,
    margin=1in
]{geometry}

\usepackage[hyphens]{url}  
\usepackage{graphicx} 
\usepackage{natbib} 
\usepackage{caption} 
\usepackage{booktabs}
\usepackage[caption=false]{subfig}
\usepackage{amsmath,amssymb,amsfonts}

\usepackage{algorithm}
\usepackage{algorithmic}

\def\BibTeX{{\rm B\kern-.05em{\sc i\kern-.025em b}\kern-.08em
    T\kern-.1667em\lower.7ex\hbox{E}\kern-.125emX}}

\DeclareMathOperator{\accuracy}{accuracy}

\title{Pruning Binarized Neural Networks: A Dedicated Framework and Globally Weighted Algorithms}

\author{
Roan Rubiales,
Jean Pierre David
}

\affiliations{
Polytechnique Montreal \\
\{roan.rubiales, jean-pierre.david\}@polymtl.ca
}
    
\begin{document}

\maketitle

\begin{abstract}

Extreme compression of deep neural networks, up to full binarization, dramatically reduces memory footprint and arithmetic complexity, facilitating deployment on constrained edge hardware with field-programmable gate arrays (FPGAs) and microcontrollers. Although combining binarization with pruning promises additional efficiency gains, existing pruning strategies are ill-suited to binarized representations and rarely translate into meaningful hardware savings. We introduce a PyTorch-based, research-oriented framework that incorporates freezing and pruning mechanisms for designing and optimizing binarized neural networks. The framework enables rapid and reproducible evaluation of state-of-the-art approaches and the fast prototyping of new ones. Leveraging this framework, we propose a novel pruning method that accounts for the relative importance of learned parameters across abstraction levels. Such a global weighting mechanism consistently achieves a superior trade-off between model accuracy and pruning rate, achieving a 70\% pruning rate on VGG11 with constant accuracy, while state-of-the-art results reach only 41\% in the binarized setting. 

\end{abstract}
\section{Introduction}
Over the past decade, deep-learning-powered technologies have found many applications across a variety of fields in computer science and even in everyday use. Deep learning has become more prevalent across our society thanks to advances in model architecture and in high-performance computing hardware, especially GPUs and AI-specific ASICs. However, deep learning and, more generally, AI require a substantial number of floating-point operations per inference, leading to high power consumption that limits deployment on edge platforms and restricts use in domains such as robotics, medical implants, and AIoT devices.


To address this challenge, the research community has developed methods to adapt state-of-the-art deep learning models to the constraints of edge platforms such as microcontrollers and FPGAs. Among the most widely adopted techniques are quantization \cite{gholami_survey_2021, courbariaux_binaryconnect_2015}, pruning \cite{liang_pruning_2021, chen_latent_2023, munagala_stq-nets_2020, li_bnn_2020} and lighter architectures \cite{sandler_mobilenetv2_2018, rastegari_xnor-net_2016, liu_reactnet_2020, zhang_fracbnn_2021, martinez_training_2020}. 

In practice, these techniques can often be combined to reduce model size and FLOPs — sometimes even eliminating floating-point operations altogether — while preserving high accuracy. Nevertheless, they remain insufficient to enable efficient FPGA or microcontroller implementations of complex deep learning models for several reasons. First, the computational complexity of multiply-accumulate (MAC) operations, especially in floating-point arithmetic, is prohibitive on FPGAs, as each floating-point operation incurs a substantial hardware cost. 
Although integer and fixed-point arithmetic significantly reduce this cost, the sheer number of operations required by most state-of-the-art architectures still exceeds what can be efficiently synthesized on an FPGA. Second, the memory footprint of the weights presents a major bottleneck. On FPGAs, each pipeline stage can access quite limited on-chip storage resources (flip-flops and embedded memories), forcing many models to rely on external memory. This reliance severely degrades performance and negates the advantages of hardware acceleration.

Among the most effective approaches for reducing both the number and complexity of operations is binarization \cite{courbariaux_binaryconnect_2015, rastegari_xnor-net_2016}, an extreme form of quantization in which all weights and activations are constrained to 1-bit precision. While not yet widely adopted, binarization minimizes the hardware cost of arithmetic operations and enables highly efficient implementations.

Pruning binarized neural networks is inherently challenging because all weights have the same magnitude. Nevertheless, pruning strategies specifically adapted to this setting can achieve significantly higher compression rates without degrading model accuracy. Prior work further demonstrates that pruning can reduce not only model size but also generalization error, acting as an implicit regularizer in binarized networks. 

Despite its effectiveness and widespread use, joint pruning and binarization remain poorly supported by mainstream deep learning frameworks, posing substantial challenges for research and development in this area. In this paper, we propose three contributions addressing these challenges:


\begin{itemize}
    \item We propose a proof-of-concept PyTorch framework that enables hardware-oriented training of binarized neural networks and their adaptation via customizable pruning and freezing (sparse training) algorithms.
    \item We formulate a global weighting mechanism for magnitude-based pruning algorithms that guides the pruning process, yielding a superior ratio of pruning rate to accuracy.
    \item Built on the two previous contributions, we propose three binary-aware pruning algorithms to optimize and convert a full-precision pretrained neural network into a binary one.
\end{itemize}

The rest of this paper is organized as follows. Section \ref{sec:rw} reviews related work on both Binary Neural Networks (BNNs) and pruning algorithms. Sections \ref{sec:contrib1} and \ref{sec:contrib2} describe our proposed framework and pruning algorithms. The test methodology and results are presented in Section \ref{sec:exp} and discussed in Section \ref{sec:disc}. The last section concludes with suggestions for future work.
\section{Related works}
\label{sec:rw}
Binary neural networks are an extreme case of quantized neural networks, in which the weights are quantized to 1 bit (2 possible values, usually $\{-1,1\}$). This idea emerged early in the history of quantization, but naive binarization often led to substantial loss of precision \cite{gholami_survey_2021}. Handling binarization differently than higher precisions became a necessity, hence the development of Binary Aware Training (BAT) \cite{courbariaux_binaryconnect_2015, courbariaux_binarized_2016}, allowing networks to be trained with real-valued weights (called the latent weights) which are dynamically converted to binary weights during the forward pass but kept in full precision for the updates. Since the binarization function is not differentiable, it is replaced by the straight-through estimator (STE) during backpropagation.  

Building on early efforts such as BAT, much of the literature focuses on designing network architectures specifically tailored to binary training. These works typically start from widely adopted backbones (e.g., MobileNetV2 \cite{sandler_mobilenetv2_2018}) and modify activation functions, building blocks, and layer ordering to better accommodate binarization \cite{rastegari_xnor-net_2016, liu_reactnet_2020, zhang_fracbnn_2021}. While Binary Neural Networks are inherently well-suited for deployment on highly resource-constrained edge devices—particularly FPGAs and microcontrollers—additional architectural and algorithmic optimizations are still required to ensure efficient and synthesizable hardware implementations \cite{ebrahimi_efficient_2023, chidambaram_poet-bin_2020}.

Pruning consists of removing a subset of a model’s weights in order to reduce its size and computational cost, while potentially improving accuracy through a regularization effect. Pruning can be applied in a structured manner, by eliminating entire neurons, channels, or kernels, or in an unstructured manner, by removing individual weights. In practice, however, only structured pruning yields a tangible reduction in computational complexity and resource usage in most deployment scenarios, as hardware is not optimized for sparse computation. Pruning may be applied in conjunction with quantization techniques \cite{liang_pruning_2021}, and is even more efficient when both are developed jointly. 

The BNN pruning landscape comprises algorithms that focus mainly on two metrics of the latent real-valued weights: their magnitude \cite{chen_latent_2023, munagala_stq-nets_2020} and the frequency of sign oscillations \cite{li_bnn_2020}. Since binarization yields weights with identical amplitudes, pruning algorithms often focus on the latent weights rather than the binarized ones. 

Magnitude-based pruning is widely adopted and has proven effective. It removes weights with small absolute values because they contribute less to driving activations beyond their thresholds than larger weights do. These methods rely on defining a pruning threshold below which weights are discarded. However, selecting a single global threshold across multiple channels is often problematic. Since weights across channels can vary by several orders of magnitude, a uniform threshold may yield suboptimal pruning decisions and reduced effectiveness. Other methods integrate pruning into the training process, using ternarization \cite{munagala_stq-nets_2020} instead of binarization. When the magnitude of the latent weight is below a threshold, the "binary" weight is set to 0 rather than the usual +1/-1.

We refer to freezing algorithms as dynamic sparse update methods, which select a sub-network within the architecture during training based on a metric, such as the cosine similarity between a channel's outputs at times $t$ and $t+1$ \cite{bragagnolo_update_2022, quelennec_towards_2024}. Sparse update is known as a technique that enables on-device training under low memory budgets \cite{lin_-device_2022}. In our setting, freezing is used as a regularization and early stopping method. Over time, more and more channels are frozen (no longer updated) until they are all frozen, and only biases are finally updated.

\section{Proposed Binary Neural Network framework}
\label{sec:contrib1}

\begin{figure}[ht]
    \centering
  \includegraphics[width=0.80\linewidth]{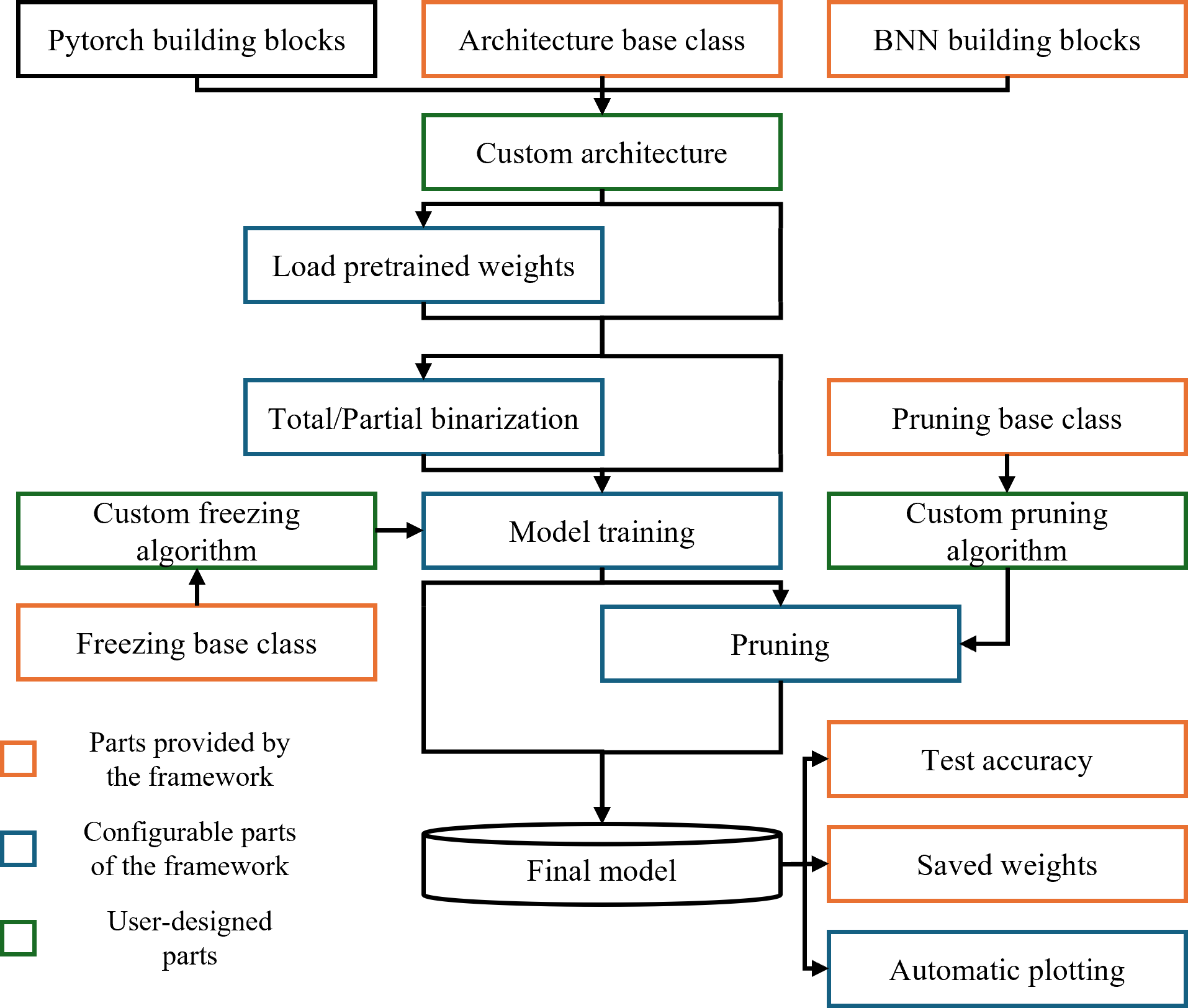}
  \caption{Block diagram describing our BNN framework.} 
  \label{fig:BDiag}
\end{figure}

Our first contribution is an experimental framework implemented in the \texttt{torch} Python library. It automates cross-experiments and facilitates comparisons of architectures, pruning, and freezing algorithms across different parameter sets.


As illustrated in Fig. \ref{fig:BDiag}, multiple aspects of the framework are directly customizable in the code: BNN architectures, pruning, and freezing algorithms. Other parts (represented by blue frames in the diagram) are configured with TOML files. These configuration files specify which parameters to use (e.g., pretrained weights or architecture hyperparameters) and are required to launch an experiment.

Our framework offers a foundational base class that all BNN modules must extend. Model architectures can then be constructed using the provided core components (e.g., binary convolutional layers, sign activation functions, etc.) or by defining custom modules derived from this base class. Standard PyTorch modules may also be integrated into the architecture; however, they will not be binarized within our framework. Pruning and freezing algorithms are also derived from a base class that provides the necessary interface functions for the algorithms to interact properly with the model and the training procedure.

We used the framework to implement several existing pruning algorithms for validation purposes. Then, we explored different variants of these algorithms to assess their impact on model accuracy and compression rate. During these experiments, we observed that normalizing the weights of each neuron or kernel independently before pruning could improve performance. In the next section, we propose several pruning algorithms derived from these experiments. 
\section{Proposed pruning algorithms}
\label{sec:contrib2}

\subsection{Global Weighting for pruning}
One of the noticeable features of our pruning algorithms is their use of transformations applied to copies of the weights to influence how the pruning thresholds are ascertained, a technique we call "global weighting". This weighting is designed to bring the channels to the same order of magnitude so that only differences between weights of the same channel (or, respectively, the same layer) remain.

The first notable transformation is \textit{Batch Normalization Folding}, which involves removing the Batch Normalization layer and integrating its weights, bias, expected value, and standard deviation into the preceding layer. In a dense layer, with $W_k$ the $k$-th neuron's weight vector and $b_k$ its bias, $\gamma_k$, $\beta_k$, $\mathbb{E}_k$, and $\sigma_k$ respectively the batch normalization layer's weight, bias, expected value and standard deviation at the $k$-th neuron, we have:

\begin{equation}
    W_k \gets \frac{\gamma_k}{\sigma_k} \cdot W_k
\end{equation}

\begin{equation}
    b_k \gets \beta_k + \frac{\gamma_k \cdot (b_k - \mathbb{E}_k)}{\sigma_k}
\end{equation}

Another transformation, which can be combined with Batch Normalization Folding, is Weight Normalization, which normalizes the weight tensors with respect to a $p$-norm (e.g., 1, 2, or infinity) along the layers or channels. Normalizing along channels cancels the batch normalization folding (on the weights), so when the batch normalization layer has been folded, we always normalize along the layers (or we do not normalize).

Once the global weighting is complete, we use the resulting weights, rather than their original counterparts, in our pruning algorithm to determine the pruning thresholds. Afterward, all global weights falling under a pruning threshold have their original counterpart pruned instead. This pruning mechanism is referred to as $\texttt{Prune\_Layers(}s,W,\bar{W}\texttt{)}$ in our algorithms, where $s$ is the threshold, $W$ is the list of weight tensors, and $\bar{W}$ is the list of globally weighted tensors. Thus, this step enables computing thresholds over global orderings of the magnitudes of the weights that are different from the naive order, allowing for higher pruning rates with similar post-pruning accuracy.

\subsection{Pruning before binarization}
A particularity of our training setup is its pretraining phase, in which only the model's activations are binarized, while the weights remain real-valued, yielding a good initialization from which the model can be fully binarized and fine-tuned afterward. We choose to prune the model at the end of pretraining, just before full binarization. We expect this unusual setup to enable the model to recover greater accuracy during fine-tuning than pruning after binarization does.

Indeed, binarization introduces noise into the weights, which is partially mitigated by fine-tuning and helps the model achieve better generalization, leading to higher test accuracy when fully binarized than with activation binarization only. We estimate that this denoising effect occurs because of the higher learning rate we start from during the post-binarization fine-tuning (see Section \ref{sec:exp}, Experimental setup, Dataset and training procedure). When the model is pruned, we introduce a second form of noise which mixes with the binarization noise. Both are mitigated during fine-tuning.

\subsection{Single threshold magnitude pruning}

\begin{algorithm}[t!]
    \caption{Single Threshold Pruning Algorithm (Search)}
    \label{alg:pruning_dicho1}
    \textbf{Input} Model's weights $W$ and a target accuracy $acc_t$. \\
    \textbf{Output} The neural network with pruned weights.

    \begin{algorithmic}[1]
    \STATE $\bar{W} \gets \texttt{Global\_Weighting(}W\texttt{)}$ 
    \STATE $\Omega \gets \texttt{Concatenate\_And\_Sort(}\bar{W}\texttt{)}$ \label{alg:pruning_dicho1:list}

    \STATE $N_\Omega \gets \texttt{length(}\Omega\texttt{)} $
    \STATE $a \gets 0$
    \STATE $b \gets N_{\Omega} - 1$
    
    \WHILE{$a < b$}
        \STATE $s \gets \frac{a+b}{2}$
        \STATE $\texttt{Prune\_Layers(}s, W, \bar{W}\texttt{)}$ \label{alg:pruning_dicho1:prune} 
        
        \IF{$\texttt{Val\_Accuracy()} > acc_t$} \label{alg:pruning_dicho1:acc_check}
            \STATE $a \gets s$ \hspace{8mm} \texttt{//keep higher thresholds} \label{alg:pruning_dicho1:higher}
        \ELSE
            \STATE $b \gets s$ \hspace{8mm} \texttt{//keep lower thresholds} \label{alg:pruning_dicho1:lower}
        \ENDIF
    \ENDWHILE
    \end{algorithmic}
\end{algorithm}

Our first algorithm (Algorithm \ref{alg:pruning_dicho1}) is a variant of the classic magnitude pruning that searches for a single threshold to prune the entire network. As mentioned earlier, a single pruning threshold may be inefficient for weights across different layers and channels. To address this issue, we use the aforementioned global weighting. A great advantage of that method is its apparent simplicity, making the single threshold directly equivalent to the pruning ratio of the model.

We then perform a dichotomous search to select the optimal threshold among the global weights. We combine the absolute value of the global weight tensors into a single list, sort them in ascending order (line \ref{alg:pruning_dicho1:list}), and then search for the pruning threshold. The midpoint is selected as the current threshold with which we prune the whole model (line \ref{alg:pruning_dicho1:prune}), and the next search interval is chosen: if the validation accuracy obtained with this threshold is suitable (line \ref{alg:pruning_dicho1:acc_check}), the upper interval is selected (leading to higher pruning rates, line \ref{alg:pruning_dicho1:higher}); otherwise, the lower interval is selected (leading to lower pruning rates, line \ref{alg:pruning_dicho1:lower}). The final threshold is found when the interval bounds meet.

\begin{algorithm}[t!]
    \caption{Single Threshold Pruning Algorithm (Sweep)}
    \label{alg:pruning_sweep}
    \textbf{Input} Model's weights $W$ and a pruning ratio step size $\alpha$. \\
    \textbf{Output} The neural network with pruned weights.

    \begin{algorithmic}[1]
    \STATE $\bar{W} \gets \texttt{Global\_Weighting(}W\texttt{)}$ 
    \STATE $\Omega \gets \texttt{Concatenate\_And\_Sort(}\bar{W}\texttt{)}$ \label{alg:pruning_sweep:list}

    \STATE $N_\Omega \gets \texttt{length(}\Omega\texttt{)} $
    \STATE $s \gets 0$
    
    \WHILE{$s < N_\Omega$}
        \STATE $\texttt{Prune\_Layers(}s, W, \bar{W}\texttt{)}$ \label{alg:pruning_sweep:prune} 
        \STATE \texttt{Collect\_Metrics()} \label{alg:pruning_sweep:collect}       
        \STATE $s \gets s + \alpha N_\Omega$
    \ENDWHILE
    \STATE \texttt{Plot\_Landscape\_Curve()} \label{alg:pruning_sweep:plot}
    \STATE $s \gets \texttt{User\_Input()}$
    \STATE $\texttt{Prune\_Layers(}s, W, \bar{W}\texttt{)}$
    \end{algorithmic}
\end{algorithm}

We also adapted the single-threshold algorithm into a sweep variant that measures validation loss and accuracy across several pruning ratios (Algorithm \ref{alg:pruning_sweep}). Instead of searching for the best pruning ratio that fits our accuracy target, we prune and compute metrics over a list of evenly spaced pruning ratios within the interval $[0\%, 100\%]$ (line \ref{alg:pruning_sweep:collect}). The landscape of accuracy (or loss) is then plotted as a function of the pruning ratio, and the selection of the final threshold is left to the assessment of the user (line \ref{alg:pruning_sweep:plot}). This variant allows for a more in-depth analysis of the pruning mechanics for each possible weighting across different pruning levels and enables extensive benchmarking of diverse global weightings.

\subsection{Accuracy-gradient ascent algorithm}

\begin{algorithm}[t!]
    \caption{Accuracy-gradient Pruning Algorithm}
    \label{alg:pruning}
    \textbf{Inputs} Model's weights $W$, an estimation interval size $h$ and number of subdivisions $m$, a target accuracy $acc_t$, a step size $\eta$. \\
    \textbf{Output} The new neural network with pruned weights.

    \begin{algorithmic}[1]
    \STATE $\bar{W} \gets \texttt{Global\_Weighting(}W\texttt{)}$ \label{alg:pruning:norm}
    \STATE $s_1, \dots, s_n \gets 0$
    \STATE $s \gets (s_1, \dots, s_n)$
    \STATE $\text{stage} \gets 1$

    \WHILE{$\texttt{Val\_Accuracy()} > acc_t$}
        \FOR{$\text{every layer } l \text{ in the network}$} 
            \STATE $\frac{\partial acc}{\partial s_l} \gets \texttt{Approx\_Derivative(}s_l,h,m\texttt{)}$ \label{alg:pruning:deriv}
        \ENDFOR
        \STATE \textbf{if} $\displaystyle\max_{l \in \{1, \dots, n\}}{\frac{\partial acc}{\partial s_l}} \leq 0$ \textbf{then} $\text{stage} \gets 2$ \label{alg:pruning:phase2}
            \IF{$\text{stage} = 1$} 
                \STATE $\texttt{Step\_Positive\_Directions(}s, \frac{\partial acc}{\partial s}, \eta\texttt{)}$ \label{alg:pruning:step1}
            \ELSE
                \STATE $\texttt{Step\_Highest\_Direction(}s, \frac{\partial acc}{\partial s}, \eta\texttt{)}$ \label{alg:pruning:step2}
            \ENDIF
            \STATE $\texttt{Prune\_Layers(}s, W, \bar{W}\texttt{)}$ \label{alg:pruning:prune}
    \ENDWHILE
    \end{algorithmic}
\end{algorithm}

The next algorithm we introduce (Algorithm \ref{alg:pruning}) uses a heuristic gradient ascent to determine the optimal pruning level for each layer given a tolerance on the target accuracy. 

As in the previous algorithms, we compute a layer's pruning mask from a channel-wise normalized weight tensor (line \ref{alg:pruning:norm}) to equalize channel importance. Both pruning and accuracy computation require selection operations, which are non-differentiable, preventing automatic differentiation. Consequently, we approximate the derivative using a finite-difference scheme.
In order to reduce the approximation noise, we perform an average finite difference (Eq. \ref{eq:f-diff}), although it requires more evaluations of the accuracy. We sample the model accuracy with the current layer pruned at different levels around the threshold $s_l$ ($acc(s_l + kh)$ and $acc(s_l - kh)$), and all the other layers pruned with their own threshold $(s_i)_{i \neq l}$ (line \ref{alg:pruning:deriv}). 

\begin{equation}
    \label{eq:f-diff}
    \frac{\partial acc}{\partial s_l} \approx \frac{1}{m} \displaystyle \sum_{k = 1}^m \frac{acc(s_l + kh) - acc(s_l - kh)}{2kh}
\end{equation}

\begin{table}[b!]
  \centering
  \begin{tabular}{lllllll}
    \toprule
    Method     & 
    \begin{tabular}[t]{@{}c@{}}Baseline\\accuracy\\(\%)\end{tabular}     & 
    \begin{tabular}[t]{@{}c@{}}Pruned\\accuracy\\(\%)\end{tabular}  &  
    \begin{tabular}[t]{@{}c@{}}Weights\\removed\\(\%)\end{tabular} \\
    \midrule
    Layer sensitivity                   & 80.6    & 80.9 & 27.0 \\
    Layer sensitivity*                  & 80.5    & 80.7 & 27.4 \\
    Layer sensitivity**                 & 89.7    & 89.6 & 32.8 \\
    Algorithm \ref{alg:pruning_dicho1}  & 89.7    & 89.3 & 46.5 \\
    Algorithm \ref{alg:pruning_sweep}   & 89.7    & 89.7 & 45.0 \\
    Algorithm \ref{alg:pruning}         & 89.7    & 89.7 & 27.9 \\
    \bottomrule
  \end{tabular}
  \caption{Comparison of pruning algorithms on the BinaryNet architecture (14M parameters).
  * Result reproduced in our environment without data augmentation (as in the original paper).
  ** Result reproduced in our environment with data augmentation.}
  \label{tab:pruning1}
\end{table}

In the first phase, we exploit the regularizing effect of thresholding by increasing the thresholds $s_l$ with positive derivatives (line \ref{alg:pruning:step1}) to move in directions that increase accuracy. A gradient ascent is applied to these thresholds (line \ref{alg:pruning:step1}), while the others are left unchanged. We use the input step size, which is usually half the size of the approximation interval, $\eta = kh$. The algorithm switches to the second phase when all directions have negative derivatives (line \ref{alg:pruning:phase2}).

In the second phase, we increase the threshold of the layer with the highest derivative, which is actually the least negative. The threshold is increased proportionally to the absolute value of the derivative and the step size (line \ref{alg:pruning:step2}). In this way, we always increase the pruning threshold that reduces accuracy the least. The algorithm terminates when the target accuracy is reached. 

When all thresholds have been shifted, the layers are pruned according to their own threshold (line \ref{alg:pruning:prune}). In this algorithm, the \texttt{Prune\_Layers} function takes a collection of thresholds as input (one for each layer).
\section{Experiments}
\label{sec:exp}

\subsection{Experimental setup}

\subsubsection{Architectures}
The main architecture on which we experimented is the binarized ConvNet architecture used by Courbariaux \textit{et al.} in their early experiments on binarization \cite{courbariaux_binaryconnect_2015, courbariaux_binarized_2016}, also referred to as BinaryNet. It consists of 6 convolutional layers and 3 dense layers and is specifically designed for small datasets such as CIFAR-10. We also used other CNN architectures (VGG11 \cite{simonyan_very_2015}, NiN \cite{lin_network_2014}) to compare our proposed methods to other state-of-the-art methods in binary CNN pruning.


\begin{table}[t!]
  \centering
  \begin{tabular}{lllllll}
    \toprule
    Method     & 
    \begin{tabular}[t]{@{}c@{}}Baseline\\accuracy\\(\%)\end{tabular}     & 
    \begin{tabular}[t]{@{}c@{}}Pruned\\accuracy\\(\%)\end{tabular}  &  
    \begin{tabular}[t]{@{}c@{}}Weights\\removed\\(\%)\end{tabular} \\
    \midrule
    Ternary pruning                     & 84.4    & 84.4 & 41.0 \\
    Ternary pruning*                    & 84.8    & 84.4 & 41.0 \\
    Algorithm \ref{alg:pruning_sweep}   & 84.8    & 84.4 & 70.0 \\
    Algorithm \ref{alg:pruning}         & 84.8    & 84.8 & 56.3 \\
    \bottomrule
  \end{tabular}
  \caption{Comparison of pruning algorithms test accuracies and pruning rates on the VGG11 architecture (28M parameters). 
  * Result reproduced in our environment.}
  \label{tab:pruning2}
\end{table}

\begin{table}[t!]
  \centering
  \begin{tabular}{lllllll}
    \toprule
    Method     & 
    \begin{tabular}[t]{@{}c@{}}Baseline\\accuracy\\(\%)\end{tabular}     & 
    \begin{tabular}[t]{@{}c@{}}Pruned\\accuracy\\(\%)\end{tabular}  &  
    \begin{tabular}[t]{@{}c@{}}Weights\\removed\\(\%)\end{tabular} \\
    \midrule
    Weight-flipping                     & 86.5    & 86.0 & 16.6 \\
    Ternary pruning                     & 85.3    & 84.1 & 39.0 \\
    Algorithm \ref{alg:pruning_sweep}   & 84.6    & 84.6 &  49.9 \\
    \bottomrule
  \end{tabular}
  \caption{Comparison of pruning algorithms on the NiN architecture (966k parameters).}
  \label{tab:pruning3}
\end{table}

\begin{figure*}
    \centering
    \subfloat[BinaryNet]{
        \includegraphics[width=0.3\linewidth]{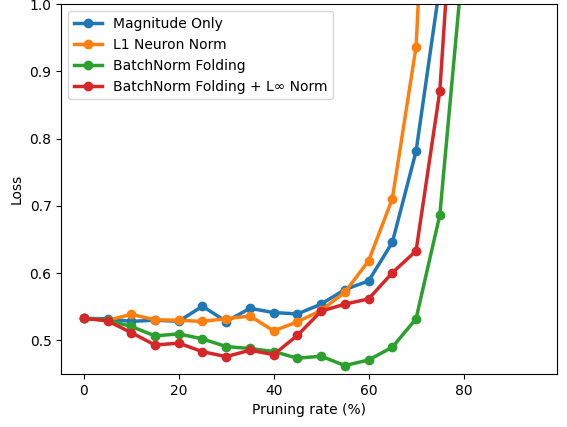}
        \label{fig:GW_comparison_bnet}
    } 
    \subfloat[VGG11]{
        \includegraphics[width=0.3\linewidth]{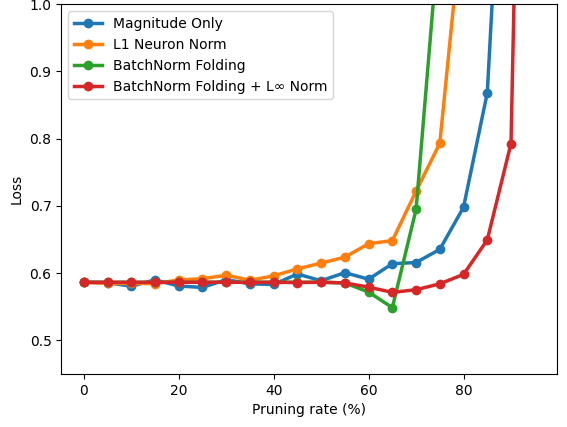}
        \label{fig:GW_comparison_vgg}
    } 
    \subfloat[NiN]{
        \includegraphics[width=0.3\linewidth]{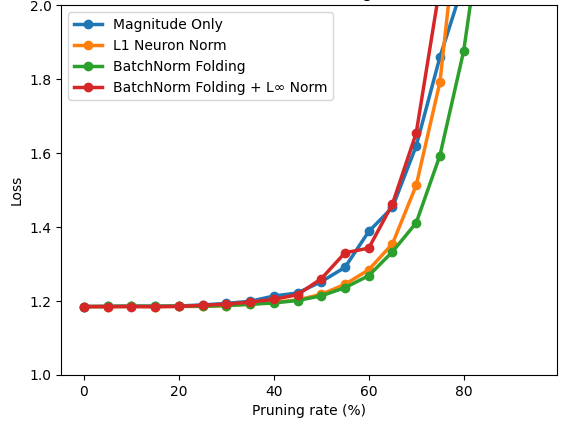}
        \label{fig:GW_comparison_nin}
    }
    \caption{Landscapes of single pruning thresholds and measured losses for several global weighting transforms, obtained through algorithm \ref{alg:pruning_sweep}.} 
    \label{fig:GW_comparison}
\end{figure*}

\subsubsection{Dataset and training procedure}
We focused our experiments on the CIFAR-10 dataset \cite{krizhevsky_learning_2009}, although our framework allows working with different datasets such as MNIST, CIFAR-100 \cite{krizhevsky_learning_2009} or ImageNet. 
We split the training set, reserving 20\% as a validation subset to monitor the training process and for early stopping via weight freezing.

We perform data augmentation (random horizontal flip and random crop) on the training set and normalize all sets using the training set's mean and standard deviation.

We use the training protocol designed by Zhang \textit{et al.} \cite{zhang_fracbnn_2021} for their CIFAR-10 BNNs, derived from the Real-To-Binary net procedure \cite{martinez_training_2020}. The training procedure is divided into two stages. The first stage is a pretraining phase, during which weights remain real-valued while activations are binarized, aiming to learn a suitable initialization for the latent weights (300 epochs). We apply our pruning algorithms after this pretraining and follow with 100 epochs of fine-tuning. The second stage is a binarization and fine-tuning phase, where both weights and activations are binarized, and lasts for 300 epochs; the latent weights are initialized using those obtained from the first phase. On the last 50 epochs of stage 2, we measure the test accuracy and return the mean as the final accuracy of the model.

For both phases, we use the Adam \cite{kingma_adam_2017} optimizer with a learning rate of $10^{-3}$ decaying 3 times every 50 epochs by $10^{-1}$ down to $10^{-6}$ (epochs 100, 150, and 200), and $(\beta_1,\beta_2) = (0.9, 0.999)$. All other parameters can be found in the configuration files for the experiments in our experimental framework GitHub repository.

Finally, we point out that model fine-tuning becomes increasingly unstable at higher pruning rates. Indeed, the model may diverge; to mitigate this risk, we apply a channel-velocity-freezing algorithm \cite{bragagnolo_update_2022} during the binary fine-tuning phase to gradually halt weight updates. It is used only when the pruned model diverges and also acts as an early stopping condition when more than 95\% of the weights are frozen.

\subsection{Pruning experiments}






\begin{table*}
 \caption{Performance of each tested global weighting transformation on BinaryNet at different pruning rates}
  \centering
  \begin{tabular}{lllllll}
    \toprule
    \begin{tabular}[t]{@{}c@{}}Target pruning\\rate (\%)\end{tabular}    &
    Method     & 
    \begin{tabular}[t]{@{}c@{}}Validation loss\\after pruning\end{tabular}    &
    \begin{tabular}[t]{@{}c@{}}Baseline test\\accuracy (\%)\end{tabular}     & 
    \begin{tabular}[t]{@{}c@{}}Pruned test\\accuracy (\%)\end{tabular} \\
    \midrule
    45.0    & Magnitude only                            & $0.539$ & $89.65$ & $89.38$         \\
            & BatchNorm folding                         & $0.473$ &         & $89.68$ \\
            & BatchNorm folding + L$\infty$ layer norm  & $0.507$ &         & $89.04$ \\
    \midrule
    55.0    & Magnitude only                            & $0.575$ & $89.65$ & $88.45$         \\
            & BatchNorm folding                         & $0.462$ &         & $89.41$ \\
            & BatchNorm folding + L$\infty$ layer norm  & $0.553$ &         & $89.05$ \\
    \midrule
    65.0    & Magnitude only                            & $0.645$ & $89.65$ & $88.96$         \\
            & BatchNorm folding                         & $0.490$ &         & $89.28$ \\
            & BatchNorm folding + L$\infty$ layer norm  & $0.600$ &         & $89.16$ \\
    \bottomrule
  \end{tabular}
  \label{tab:binarynet_pruning_rebalancers}
\end{table*}

\subsubsection{Comparison of global weighting transformations}

To analyze the effect of global weighting, we first used our single-threshold sweep algorithm  with different transformations (Algorithm \ref{alg:pruning_sweep}). 

In Figure \ref{fig:GW_comparison_bnet}, we compare different weighting transformations on BinaryNet with the single-threshold pruning algorithm in its sweep variant. We notice a clear decrease in validation loss on the green curve associated with BatchNorm folding, especially when compared to the baseline represented by the blue curve. The ability of weighting transformations to minimize validation loss is key to improving pruning algorithms, especially at the high pruning rates necessary for computationally constrained inference platforms. This observation can be repeated when pruning VGG11 and NiN (Figures \ref{fig:GW_comparison_vgg}, \ref{fig:GW_comparison_nin}) in the same manner; however, the margin of optimization is thinner.

These results hold through binarization and fine-tuning (see Table \ref{tab:binarynet_pruning_rebalancers}). Indeed, global weighting yields results superior to those obtained through magnitude only (no transformation), for each tested pruning rate. We also notice that global weighting allows for better pruning-accuracy tradeoffs at rates as high as 65\% for a 0.37\% decrease in accuracy using Batch Normalization Folding. 

\subsubsection{Comparison of different pruning methods}
We compare our method's best results to various works from the literature, specifically Layer Sensitivity \cite{chen_latent_2023} (relabeled from Latent Weight pruning, since our methods leverage latent weights as well), Ternary Weights \cite{munagala_stq-nets_2020}, and Weight-flipping \cite{li_bnn_2020}. We present these comparisons in Tables \ref{tab:pruning1}, \ref{tab:pruning2} and \ref{tab:pruning3}. They report the test accuracies before and after pruning (and fine-tuning), as well as the proportion of weights removed and the original number of weights in each architecture. The final state of the model targeted by our algorithms and pruning hyperparameters attempts to match as closely as possible the accuracy delta observed in the literature while improving the pruning ratio as much as possible.

First, we reimplemented several pruning methods from the literature, yielding results that are quasi-identical to those reported in the original papers. In Table \ref{tab:pruning1}, lines 2-3, we reproduced Chen \textit{et al.}'s Layer Sensitivity with and without data augmentation, since our training procedure uses it and theirs does not. In Table \ref{tab:pruning2}, line 1, we also reproduced Munagala \textit{et al.}'s method on VGG11 with very close results. This validates our framework as a reasonable benchmarking tool for pruning. Using the proposed algorithms based on magnitude pruning, we achieve better overall results than those reported in the literature. The single threshold sweep method maintains an accuracy of 89.7\% on BinaryNet while offering a better pruning ratio of 45.0\% (Table \ref{tab:pruning1}, line 5), $1.66$ times as high as Chen \textit{et al.}'s layer sensitivity method. Our accuracy-gradient method also maintains the baseline accuracy, but with a lower 27.9\% pruning rate (Table \ref{tab:pruning1}, line 6).
On VGG11, all our algorithms achieve better pruning rates than Munagala \textit{et al.}'s STQ-B (Ternary pruning) method, with almost constant accuracy, such as a $70.0\%$ pruning ratio for our algorithm \ref{alg:pruning_sweep} (Table \ref{tab:pruning2}).
Lastly, we experimented on the NiN architecture. This architecture is much more compact than BinaryNet and VGG11 (less than 1M parameters, compared to more than 10M for the other two), which foreshadows lower pruning rates. Nevertheless, our single threshold sweep removes $49.9\%$ of the weights, at a constant accuracy of $84.6\%$ (Table \ref{tab:pruning3}).

Our Algorithm \ref{alg:pruning_sweep} used BatchNorm folding + L$\infty$ layer norm for VGG11 and BatchNorm folding for BinaryNet and NiN as its global weighting. Algorithms \ref{alg:pruning_dicho1} and \ref{alg:pruning} used L1 neuron/channel-wise norm in all cases.

\subsection{Estimated hardware gains}

As stated before, our work aims at reducing the memory and operation cost of our models on constrained platforms, namely microcontrollers and FPGAs. Our framework does not provide the necessary inference-ready implementation for such devices, but it yields reports on minimal on-device model size (without accounting for sparsity), and number of binary multiply accumulate operations (BMAC) before and after pruning. The results for our models pruned by algorithm \ref{alg:pruning_sweep} are displayed in Table \ref{tab:bmacs}, as they perform better overall.

The model size depends on the number of weights unpruned and on the precision of the weights. The binary weights are gathered in groups of 8 bits (1 byte) to account for the need for contiguity of most random access memories and storage hardware. Biases are considered to be quantized to 8 bits, assuming they can be quantized without losing precision. Equation \ref{eq:layer_size} gives the size (in bytes) of a layer depending on its unpruned weights ($N_\text{unpruned}$), biases ($N_\text{biases}$), and their precision:

\begin{table}[t!]
 \caption{Performance metrics of our models on their best result in pruning rate while keeping accuracy constant.}
  \centering
  \begin{tabular}{lllllll}
    \toprule
    Model    &
    \begin{tabular}[t]{@{}c@{}}Base\\size\end{tabular}    &
    \begin{tabular}[t]{@{}c@{}}Pruned\\size\end{tabular}    &
    \begin{tabular}[t]{@{}c@{}}Base\\BMACs \end{tabular}     &
    \begin{tabular}[t]{@{}c@{}}Pruned\\BMACs \end{tabular}  \\   
    \midrule
    BinaryNet   & 1.76MB & 0.97MB & 617M & 479M \\ 
    \midrule
    VGG11       & 3.53MB & 1.08MB & 172M & 122M \\ 
    \midrule
    NiN         & 0.18MB & 0.12MB & 222M & 127M \\ 
    \bottomrule
  \end{tabular}
  \label{tab:bmacs}
\end{table}

\begin{equation}
\label{eq:layer_size}
    S = \begin{cases}
            \lceil \frac{N_\text{unpruned}}{8} \rceil + N_\text{biases} \text{ if binarized} \\
            4 (N_\text{unpruned} + N_\text{biases}) \text{ if full precision}
        \end{cases}
\end{equation}

To assess the number of BMACs of our pruned models, we use $\text{BMAC}_\text{conv} = (W_\text{out} \times H_\text{out}) \times N_\text{unpruned}$ where $W_\text{out}$ and $H_\text{out}$ are the dimensions of the output of the layer, in the case of a convolution layer, and $\text{BMAC}_\text{FC} = N_\text{unpruned}$ in the case of a fully connected layer. Indeed, a weight accounts for more operations in a convolution layer than in a fully connected one.

\begin{figure}[t!]
    \centering
    \includegraphics[width=0.85\linewidth]{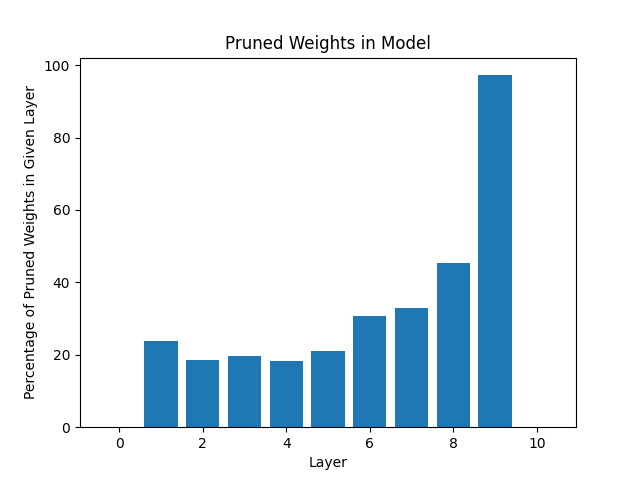}
    \caption{Pruning rates of each layer in VGG11 pruned by single threshold sweep (see Table \ref{tab:pruning2}).}
    \label{fig:pr_per_layer}
\end{figure}

As seen in Table \ref{tab:bmacs}, a reduction in size does not translate into the same reduction in the number of operations. For instance, VGG11 becomes $3.3\times$ smaller, but its maximum speedup is only $1.4\times$. This can be explained by the distribution of pruned weights across the layers (see Figure \ref{fig:pr_per_layer}), where we notice that most of the weights are pruned in fully connected layers (layers 8, 9, and 10), making the model size shrink faster than the operational cost.
\section{Discussion}
\label{sec:disc}
Previous work has demonstrated the relevance of pruning to reduce model size with minimal impact on accuracy. Nevertheless, pruning is still not integrated into the mainstream AI frameworks. We presented a proof of concept for a self-consistent AI framework that supports both pruning and binarization. Although the framework is still under development, it has already enabled systematic benchmarking of multiple pruning strategies and achieved pruning results that surpass the state of the art for several models. These outcomes highlight the importance of such an environment, particularly for edge deployments on resource-constrained platforms such as microcontrollers and FPGAs. 

Regarding our proposed pruning methods, they achieve better results than other methods, especially our single-threshold algorithms — a counter-intuitive result considering their relative simplicity. These positive results demonstrate the value of the global weighting mechanism and of pre-binarization pruning. However, they prune models in an unstructured manner (by pruning weights individually), yielding better pruning tradeoffs but with increased sparsity.

We displayed the cost of our models in binary MACs, which differ from floating-point MACs in that they are easier to optimize on most architectures. For instance, CPUs can use XNOR-Count operations as stated by Rastegari \textit{et al.} \cite{rastegari_xnor-net_2016}, while FPGAs can pipeline each operation and carry them out individually. However, the highly sparse models produced by our pruning algorithms render some of these optimizations less efficient in a manner that is difficult to estimate.

\section{Conclusions and future work}
\label{sec:conclu}

In this paper, we proposed a preliminary framework for binarization, pruning, and freezing of deep learning architectures, in order to prove its relevance and feasibility. It provides the foundation for implementing binarized architectures and pruning or freezing algorithms. To validate the approach, we reproduced a few state-of-the-art compact binarized CNNs and several popular pruning methods for BNNs. We hope similar functionalities will be adopted by the main deep learning frameworks.
Furthermore, we proposed three novel pruning methods based on an accuracy-gradient ascent that yield pruning rates up to twice those reported in the literature without compromising model precision. Although our pruning methods yield better results in terms of model size, they would greatly benefit from being converted to a structured pruning approach and from a more in-depth ablation study to ascertain the real effect of global weighting and pre-binarization pruning. Our framework would also benefit from an automatic code generation tool that provides an on-device implementation of the models pruned by our methods, completing the deployment pipeline. This is left for future work.


\bibliography{src/references}

\end{document}